\documentclass[11pt,a4paper]{article}
\usepackage[hyperref]{emnlp2020}
\usepackage{times}
\usepackage{latexsym}

\usepackage{microtype}

\usepackage{amsmath}
\usepackage{bm}
\usepackage{multirow}
\usepackage{graphicx}

\usepackage{hyperref}
\usepackage{url}
\usepackage{caption,subcaption}
\usepackage{float}
\usepackage{tabularx}
\usepackage{chngcntr}
\usepackage{xcolor}
\usepackage{enumitem}

\usepackage{multicol,lipsum}

\aclfinalcopy % Uncomment this line for the final submission
\title{Word Embedding Visualization Via Dictionary Learning}

\author{
\setlength{\tabcolsep}{10pt}
\begin{tabular}{@{}cccc@{}}
Juexiao Zhang\thanks{\ equal contribution. Correspondence to: Juexiao Zhang $<$jiaoxiao16@mails.tsinghua.edu.cn$>$, Yubei Chen $<$yubeic@eecs.berkeley.edu$>$} $^{\ 1,2}$ &  
Yubei Chen$^{*\ 1,3}$ &
Brian Cheung$^{1,3}$ & 
Bruno A Olshausen$^{1,3,4}$\\
\end{tabular}\\[5pt]
\small\ $^{1}$ Berkeley AI Research (BAIR), UC Berkeley\\
\small\ $^{2}$ Department of Electrical Engineering, Tsinghua University, Beijing, China\\
\small\ $^{3}$ Redwood Center for Theoretical Neuroscience, UC Berkeley\\
\small\ $^{4}$ Helen Wills Neuroscience Institute \& School of Optometry, UC Berkeley\\
}

\date{}
\hypersetup{draft}
\begin{document}
\maketitle
\begin{abstract}
Co-occurrence statistics based word embedding techniques have proved to be very useful in extracting the semantic and syntactic representation of words as low dimensional continuous vectors. In this work, we discovered that dictionary learning can open up these word vectors as a linear combination of more elementary word factors. We demonstrate many of the learned factors have surprisingly strong semantic or syntactic meaning corresponding to the factors previously identified manually by human inspection. Thus dictionary learning provides a powerful visualization tool for understanding word embedding representations. Furthermore, we show that the word factors can help in identifying key semantic and syntactic differences in word analogy tasks and improve upon the state-of-the-art word embedding techniques in these tasks by a large margin.
\end{abstract}

\section{Introduction}
\label{sec:intro}

Several recent works \cite{mikolov2013efficient, pennington2014glove, bojanowski2017enriching, peters2018deep} show that co-occurrence statistics can be used to efficiently learn high-quality continuous vector representations of words from a large corpus of text. The learned word vectors encode rich semantic and syntactic information, which is demonstrated with word analogies. As shown by \cite{mikolov2013linguistic}, vector arithmetic operations such as: `king' - `queen' = `man' - `woman' reflects the semantic analogy of what `king' is to `queen' as what `man' is to `woman'.  Thanks to the competitive performance of these models, these methods have become fundamental tools used in the study of natural language processing.

% \textcolor{red}{what}
% \sout{\textcolor{red}{and computational linguistics}}. 

\citet{allen2019analogies, ethayarajh2018towards, levy2014linguistic} explain and understand these word embeddings from a theoretical perspective. 
Empirically, visualizing these embeddings can be difficult since individual dimensions of most word embeddings are not semantically meaningful. % BC: you want to expand more on this, right now it seems like a contradiction to say the whole is vector meaningful even though the parts (dimensions are not). 

In the absence of tools to visualize and gain further insight about the learned word vectors, we have little hope of improving the existing performance on downstream tasks like word analogy \cite{mikolov2013efficient}. There are three major ways for visualizing word vectors: 
\begin{itemize}[leftmargin=*]
    \item Nearest neighbor approach: we can use either Euclidean distance or cosine similarity to search for each word vector's nearest neighbors to find its relevant words \cite{peters2018deep,bojanowski2017enriching,Wieting2015TowardsUP}. This method only provides a single scalar number of relatedness information while two words may exhibit much more intricate relationships than just a relatedness \cite{pennington2014glove}. For example, {\it man} and {\it woman} both describe human beings and yet are usually considered opposite in gender.
    \item{t-SNE approach \cite{maaten2008visualizing,hinton2003stochastic}: This approach nonlinearly reduces word vectors to a very low dimensional (most likely 2) space. While such a global method reveals some interesting separation between word groups, it often distorts \cite{liu2017visual} important word vector linear structures and does not exhibit more delicate components in each word.}
    % \textcolor{orange}{(BC: include citations here to give evidence this is a major visualization method)}
    \item Subset PCA approach \cite{Mikolov2013DistributedRO}: 1) Select a subset of word pairs, which have certain relations, e.g. city-country, currency-country, comparative etc. 2) Perform PCA on the selected subset and visualize the subset with the first two principle components, as shown in Figure~\ref{fig: pca-profession}. The relationship is frequently encoded by a vector in this subspace. However, performing PCA with all the word vectors makes this information entirely opaque. This method needs manually selected sets of words which requires human intervention. Despite this, such a visualization method does capture important semantic meaning for word vectors.
    % This visualization shows there are actually some interesting directions in the word vector space, which capture important semantic meanings. 
\end{itemize}

The linear substructures generated using subset PCA approach and semantic content of arithmetic operations provide strong motivation to automatically discover the factors which these underlying word vectors are composed of.

The key insight of this work is that the relationships visualized in the human selected subsets represent more elementary word factors and a word vector is a linear combination of a sparse subset of these factors. Dictionary learning \cite{olshausen1997sparse,olshausen1996emergence,bell1995information} is a useful tool to extract elementary factors from different modalities. \cite{murphy2012learning, fyshe2014interpretable, faruqui-etal-2015-sparse, subramanian2018spine} shows sparsity help to improve the dimension interpretability. Specifically, in \cite{faruqui-etal-2015-sparse}, the authors apply non-negative sparse coding (NSC) \cite{hoyer2002non, lee1999learning} with binary coefficients to word vectors and suggest to use the resulted sparse vectors as word vector representation of words. Then \cite{subramanian2018spine} followed the idea and applied k-sparse autoencoder to further improve the sparse word vector representation. In \cite{arora2018linear}, the authors also decomposed word vectors using k-sparse sparse coding and reported the discovery of ``discourse atoms", supported by a random walk model. In this work, we share the same insight of the linear structure in the word embedding space, but thoroughly explore this idea from a visualization perspective. Since a word vector may involve a different number of factors with a different strength, neither binary coefficients or a k-sparse setting would be ideal for such a purpose. In section~\ref{sec:learning} we reformulate the NSC problem to be more flexible. Further, we introduce the spectral clustering algorithm to handle group sparsity. Once NSC has been trained on word vectors from different word embedding methods, in section~\ref{sec:visual}, we demonstrate reliable word factors with very clear semantic meanings, which is consistent with but not limited to the existing prior knowledge. With these reliable word factors, we then open up the word vectors and visualize them in many different ways. Through these visualizations, we show many interesting compositions like the following:
\begin{align*}
% \vspace{-0.2in}
    \text{apple} =& 0.09``\text{dessert}" + 0.11``\text{organism}" + 0.16\\ ``\text{fruit}" 
    & + 0.22``\text{mobile\&IT}" + 0.42``\text{other}"
\end{align*}

Many new word analogy tasks can be easily developed based on these learned word factors. Different embedding models and text corpus bias can be diagnosed, i.e. we find that a factor proportion might change depend on which corpus we use to train the word vector embedding. 
% \textcolor{red}{(AL:may better provide a brief explanation on what a diagnosis is?)}.

Since several learned word factors may encode a similar meaning and frequently work together, we can use spectral clustering \cite{ng2002spectral, von2007tutorial} to group word factors with similar meanings. Each group can provide robust semantic meaning for each factor and identify key semantic differences in a word analogy task. We show in Section~\ref{sec:improve} that these groups help to improve the word analogy tasks significantly in almost every subcategory irrespective to which word embedding technique we use. Our simple and reliable discovery provides a new venue to understand the elementary factors in existing word embedding models. In Section~\ref{sec:disscuss}, we discuss a few interesting questions and point out some potential directions for future work.

\section{Word Factors Learning and Spectral Grouping}
\label{sec:learning}
We use non-negative {\em overcomplete} sparse coding to learn word factors from word vectors. Given a set of word vectors $\{x_j\in{\rm I\!R}^{n}\}$ ($n=300$ is used in this work as a convention), we assume each of them is a sparse and linear superposition of word factors $\phi_i$:
\begin{equation}\label{eqn:sparse_code}
        \bm{x} = \Phi\, \bm{\alpha} + \bm{\epsilon},\ \text{s.t.}\ \bm{\alpha} \succeq 0,
\end{equation}
where $\Phi\in{\rm I\!R}^{n\times d}$ is a matrix with columns $\Phi_{:,i}$, $\bm{\alpha}\in{\rm I\!R}^d$ is a sparse vector of coefficients to be inferred and $\bm{\epsilon}$ is a vector containing independent Gaussian noise samples, which are assumed to be small relative to $\bm{x}$.  Typically $d>n$ so that the representation is {\em overcomplete}. This inverse problem can be efficiently solved by FISTA algorithm \cite{beck2009fast}. A word vector $\bm{x}_i$ is sampled with respect to the responding frequency $f_i$ of the word $i$ in the corpus. Once the sparse coding inference is solved, we can then update the word factors to better reconstruct the word vectors. Through this iterative optimization, the word factors can be learned. We provide more details of the algorithm in Appendix~\ref{sec:optimization}.

Though overcomplete sparse coding tends to extract more accurate elementary factors and thus approximate signal vectors at better accuracy given a fixed sparsity level, several learned factors may be corresponding to a similar semantic meanings.
%BC: This sentence doesn't make much sense, more accurate compared to what? YB:Updated, please check if it's okay now.
\cite{chen2018sparse} proposes to model the relationships by using a manifold embedding, which is essentially a spectral method. In this work, we use spectral clustering \cite{ng2002spectral} to group the learned word factors into groups.

Since word factors with similar semantic meaning tends to co-activate to decompose a word vector, we calculate a normalized covariance matrix $W$ of word factor coefficients with the unit diagonal removed:
\begin{equation}\label{eqn:covariance}
        W = \sum_i{f_i \hat{\bm{\alpha}}_i\hat{\bm{\alpha}}_i^T} - I
\end{equation}
% \begin{equation}\label{eqn:covariance}
%         W = \left[\sum_i{f_i (\bm{\alpha}_i \circ \bm{\sigma})(\bm{\alpha}_i \circ \bm{\sigma}})^T\right] - I
% \end{equation}

where $f_i$ is the frequency of the $i$th word in the corpus, $\hat{\bm{\alpha}}_i$ is the normalized sparse coefficients by each dimensions standard deviation such that $\hat{\alpha}_{ij}=\alpha_{ij}/ \sigma_j$, and $\sigma_j = \left[\sum_i{f_i (\alpha_{ij})^2}\right]^{\frac{1}{2}}, \ j=1\dots n$.
% This matrix is used as the weight adjacency matrix in the spectral clustering algorithm \cite{ng2002spectral, von2007tutorial} and it captures the similarity between the word factors in a given corpus. We also leave details of the implementation in the appendix.

This matrix captures the similarity between the word factors. To better perform a spectral clustering, we first make the normalized covariance matrix $W$ sparse by selecting q largest values in each row of $W$:
\begin{equation}
    W_{sp} = F_q(W, dim=0)
\end{equation}
Where $F_q$ stands for keeping the q largest values unchanged in the given dimension, while setting all the other entities to 0. Then we obtain a symmetric sparse matrix:
\begin{equation}
    W_{adj} = W_{sp}+W_{sp}^T
\end{equation}
$W_{adj} \in {\rm I\!R}^{d\times d}$ is the  adjacency matrix used in spectral clustering \cite{ng2002spectral,von2007tutorial}.
% We use the implementation of the algorithm in Scipy \cite{scipy}.
Using the notation from \cite{von2007tutorial}, we first compute a normalized Laplacian matrix $L_{sym}$:
\begin{equation}
    L_{sym} = I - D^{-1/2}W_{adj}D^{-1/2}
\end{equation}
where $D$ is a diagonal matrix with the sum of each row of the symmetric $W_{adj}$ on its diagonal. Suppose we set the number of clusters to k, then the first k eigenvectors of $L_{sym}$ form the columns of matrix $V \in {\rm I\!R^{d \times k}}$:
\begin{equation}
    V = [v_1, v_2, ..., v_k] 
\end{equation}
And we normalize each row of V to get a new matrix $U$:
\begin{equation}
    U = [u_1, u_2, ..., u_d]^T
\end{equation}
where each row $u_i = V_{i,:}/ \Vert V_{i,:} \Vert_2$. $\Vert V_{i,:} \Vert_2$ indicates the L2 norm of the \textit{i}th row of V. Finally, a k-means algorithm is performed on the $d$ rows of U to get the clusters. 

% $\hat{\bm{\alpha}}_i=\bm{\alpha}_i \circ \bm{\sigma}$, $\circ$ denotes the the Hadamard product and $\bm{\sigma} = \left[\sum_i{f_i (\bm{\alpha}_i \circ \bm{\alpha}_i)^2}\right]^{\frac{1}{2}}$.
%BC: f() needs to be defined here, also I think you will want to decompose eqn:covariance into a couple parts. Specifically one part should explain what \alpha \circ \sigma means since it's used twice.

\section{Visualization}
\label{sec:visual}

The word factors learned for different word embedding models are qualitatively similar. % BC: needs more justification here, what makes them qualitatively similar?
For simplicity, we show the results for the 300 dimensional GloVe word vectors\cite{pennington2014glove} pretrained on CommonCrawl. We shall discuss the difference across different embedding models at the end in this section.

Once word factors have been learned and each word vector's sparse decomposition $\bm{\alpha}$ has been inferred by solving Equation~\ref{eqn:sparse_code}, we can denote all the decomposition of the word vectors in the matrix form as the following:
\begin{equation}\label{eqn:sparse_code_matrix}
        X \approx \Phi A,\ \text{s.t.}\ A \succeq 0
\end{equation}
where $X_{:,i} = \bm{x}_i$ and $A_{:,i} = \bm{\alpha}_i$, $X\in{\rm I\!R}^{n\times N}$ and $A\in{\rm I\!R}^{d\times N}$. $N$ is the size of the vocabulary. Note that dictionary $\Phi$ is not non-negative.

% We can visualize a {\it word factor} $\Phi_{:,j}$ by examining its corresponding row $A_{j,:}$ in $A$ and visualize a word vector $x_{:,i}$ by examining the corresponding column $\alpha_i$ of $A$. Since word factors are learned in an unsupervised fashion, the explicit meaning of each factor is unknown in advance. To help understand the meaning of a specific factor, we print out the words that have high coefficients for this factor, some examples are illustrated in Table \ref{tab:factor_words}. 
In the following, We first present the visualization of word factors and give each factor a semantic name. Then we decompose word vectors into linear combinations of word factors. Furthermore, we demonstrate factor groups and discuss the difference across multiple word embedding models.

% \textcolor{red}{(feel decompose is more accurate here)}
% \todo[inline]{Relocate the following}
% In order to make a clear visualization and offer a comparison with PCA visualization, we also make use of subsets of words. But note that they do not affect the factors themselves.

\subsection{Word Factor Visualization}
\label{sec:visual_wordfactor}

%Word factors can be visualized through the sparse coding coefficients of each word vector.
We can visualize a {\it word factor} $\Phi_{:,j}$ by examining its corresponding row $A_{j,:}$ in $A$ and visualize a word vector $x_{:,i}$ by examining the corresponding column $\alpha_i$ of $A$. Since word factors are learned in an unsupervised fashion, the explicit meaning of each factor is unknown in advance. To help understand the meaning of a specific factor, we print out the words that have large coefficients for this factor, some examples are illustrated in Table \ref{tab:factor_words}.
We refer to these coefficients as activations as they describe how much a factor is turned on for a specific word. 
%In Table~\ref{tab:factor_words}, we demonstrate a set of factors with their top-activation words.
Usually the top-activation words for each factor share an obvious semantic or syntactic meaning. Based on the top-activation words, a semantic name can be given to each factor as a guide. For example, for factor 59, we can call it ``medical'' since most of the words have activation on it are related to medical purpose. In the Appendix~\ref{sec:naming}, we discuss this naming procedure in more detail.

% in word factor visualization section
\begin{table*}[h]
    \centering
    \footnotesize
    \begin{tabular}{|c|l|}
\hline
\multicolumn{1}{|c|}{factors} & \multicolumn{1}{c|}{top activation words}                \\ \hline
\multirow{2}{*}{\begin{tabular}[c]{@{}c@{}}59\\ ``medical''\end{tabular}}
& hospital, medical, physician, physicians, nurse, doctor, hospitals, doctors, nurses, patient,\\ & nursing, medicine, care, healthcare, psychiatric, clinic, psychiatry, ambulance, pediatric\\ \hline
\multirow{2}{*}{\begin{tabular}[c]{@{}c@{}}116\\ ``vehicle''\end{tabular}}
& vehicles, vehicle, driving, drivers, cars, car, driver, buses, truck, trucks, taxi, parked,\\ & automobile, fleet, bus, taxis, passenger, van, automobiles, accidents, motorcycle, mph\\ \hline
\multirow{2}{*}{\begin{tabular}[c]{@{}c@{}}193\\ ``ware''\end{tabular}}
& pottery, bowl, bowls, porcelain, ware, vase, teapot, china, saucer, denby, vases, saucers,\\ & ceramic, glass, plates, earthenware, pitcher, wedgwood, pots, plate, tureen, jug, pot, jar\\ \hline
\multirow{2}{*}{\begin{tabular}[c]{@{}c@{}}296\\ ``mobile\&IT''\end{tabular}}%Apple.Inc
& ipad, iphone, ios, itunes, apple, android, app, ipod, airplay, 3g, 4s, apps, ipads, htc, tablet\\ & galaxy, jailbreak, iphones, netflix, mac, os, touch, nook, skyfire, dock, siri, eris, 4g, tablets\\ \hline
\multirow{2}{*}{\begin{tabular}[c]{@{}c@{}}337\\``superlative''\end{tabular}}
& strongest, funniest, largest, longest, oldest, fastest, wettest, tallest, heaviest, driest, sexiest,\\ & scariest, coldest, hardest, richest, biggest, happiest, smallest, toughest, warmest, most \\ \hline
\multirow{2}{*}{\begin{tabular}[c]{@{}c@{}}461\\ ``country''\end{tabular}}
& venezuela, germany, paraguay, uruguay, norway, russia, lithuania, ecuador, netherlands,\\ & estonia, korea, brazil, argentina, albania, denmark, poland, europe, sweden, colombia \\ \hline
\multirow{2}{*}{\begin{tabular}[c]{@{}c@{}}470\\ ``bedding''\end{tabular}}
& mattress, pillow, bed, mattresses, beds, pillows, queen, ottoman, simmons, cushion, bedding,\\ &topper, foam, plush, sleeper, sofa, comforter, couch, futon, seat, bolster, pad, sleeping\\ \hline
\multirow{2}{*}{\begin{tabular}[c]{@{}c@{}}493\\ ``royal''\end{tabular}}
& king, royal, throne, prince, monarch, emperor, duke, queen, reign, coronation, kings, empress,\\ & regent, dynasty, palace, monarchs, ruler, crown, heir, monarchy, kingdom, sultan, consort\\\hline
\multirow{2}{*}{\begin{tabular}[c]{@{}c@{}}582\\ ``fruit''\end{tabular}}
& fruit, fruits, pears, oranges, apples, peaches, grapes, apple, ripe, plums, bananas, mandarin,\\ & grapefruit, peach, berries, tomatoes, kiwi, watermelon, berry, lemons, mango, canning, kiwis\\ \hline
\multirow{2}{*}{\begin{tabular}[c]{@{}c@{}}635\\ ``Chinese''\end{tabular}}
& china, fujian, zhejiang, guangdong, hangzhou, shandong, shanghai, qingdao, beijing,\\ & chongqing, guangzhou, sichuan, jiangsu, hainan, hebei, luoyang, shenzhen, nanjing, henan\\ \hline
\multirow{2}{*}{\begin{tabular}[c]{@{}c@{}}781\\ ``national''\end{tabular}}
& croatian, american, lithuanian, norwegian, vietnamese, chinese, romanian, bulgarian,\\ & indonesian, armenian serbian, turkish, hungarian, korean, malaysian, italian, austrian \\ \hline
\multirow{2}{*}{\begin{tabular}[c]{@{}c@{}}886\\ ``female''\end{tabular}}
& her, queen, herself, she, actress, feminist, heroine, princess, empress, sisters, woman, dowager,\\ & lady, sister, mother, goddess, women, daughter, diva, maiden, girl, née, feminism, heroines\\ \hline
\end{tabular}
    \caption{In this figure we show a set of learned factors with its top-activation words. Based on the common aspect of the first 20\% of the top-activation words (usually around 100 words), we can give each of the factors a semantic name.}
    \label{tab:factor_words}
\end{table*}

\begin{figure}[h]
\begin{center}
\includegraphics[width=0.8\linewidth]{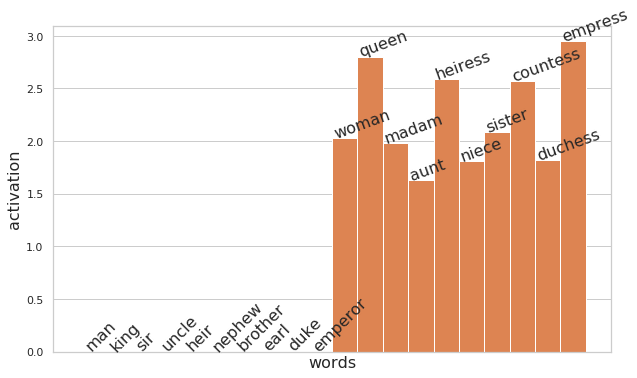}
\end{center}
\caption{``Female'' factor's activation w.r.t. a selected set of words contain both male-related words and female related words.}
\label{fig: genderbar}
% \vspace{-0.1in}
\end{figure}

\begin{figure}[htb]
\begin{center}
\includegraphics[width=0.8\linewidth]{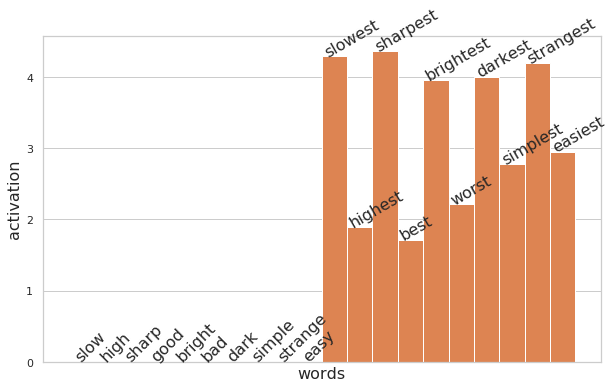}
\end{center}
\caption{``Superlative'' factor's activation w.r.t. a selected set of words contain words and their superlative forms.}
\label{fig: superlativebar}
\end{figure}

\noindent {\bf Reliability.}
The factors we discovered exhibit strong reliability. For instance, a female factor is illustrated in Figure \ref{fig: genderbar}. Clearly, the activations remain all 0s for the male words, but have high values for the female words. Similarly strong word factors are also found to capture syntactic meanings of words. Figure \ref{fig: superlativebar} shows activations on a factor representing the superlative information, i.e. the superlative factor, where the superlative forms of adjectives have relatively large coefficients. The significance is obvious from the sharp contrast in the heights of the bars. 

\begin{figure}[h]
\vspace{-0.2in}
\begin{center}
\includegraphics[width=0.8\linewidth]{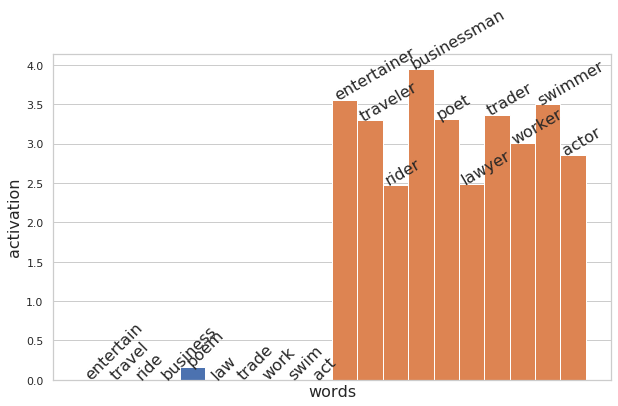}
\end{center}
\caption{``Profession'' factor's activation w.r.t. a selected set of words contain action-related words and their profession form.}
\label{fig: professionbar}
\end{figure}

\noindent {\bf The Learned Factors Match the Prior.}
We empirically find that for each of the 14 word analogy tasks, there are always a few corresponding factors capture the key semantic difference, e.g. the ``female'', ``superlative'', ``country'' factors shown in Table~\ref{tab:factor_words}. Given the learned word factors closely matched 
the 14 word analogy tasks chosen based on human priors, we can expect the rest majority of the learned factors may provide an automatic method to select and construct the word analogy tasks. For instance, Figure \ref{fig: professionbar} shows a factor corresponding to professions. For words such as ``entertain'', ``poem'', ``law'' and so on, it has 0 or very small activation, whereas for ``entertainer'', ``poet'' and ``lawyer'', the activations are clearly large. In Appendix~\ref{sec:newanalogy}, we provide more of such generated word analogy tasks by the learned factors.

\begin{figure}[h!]
\begin{center}
\includegraphics[width=0.8\linewidth]{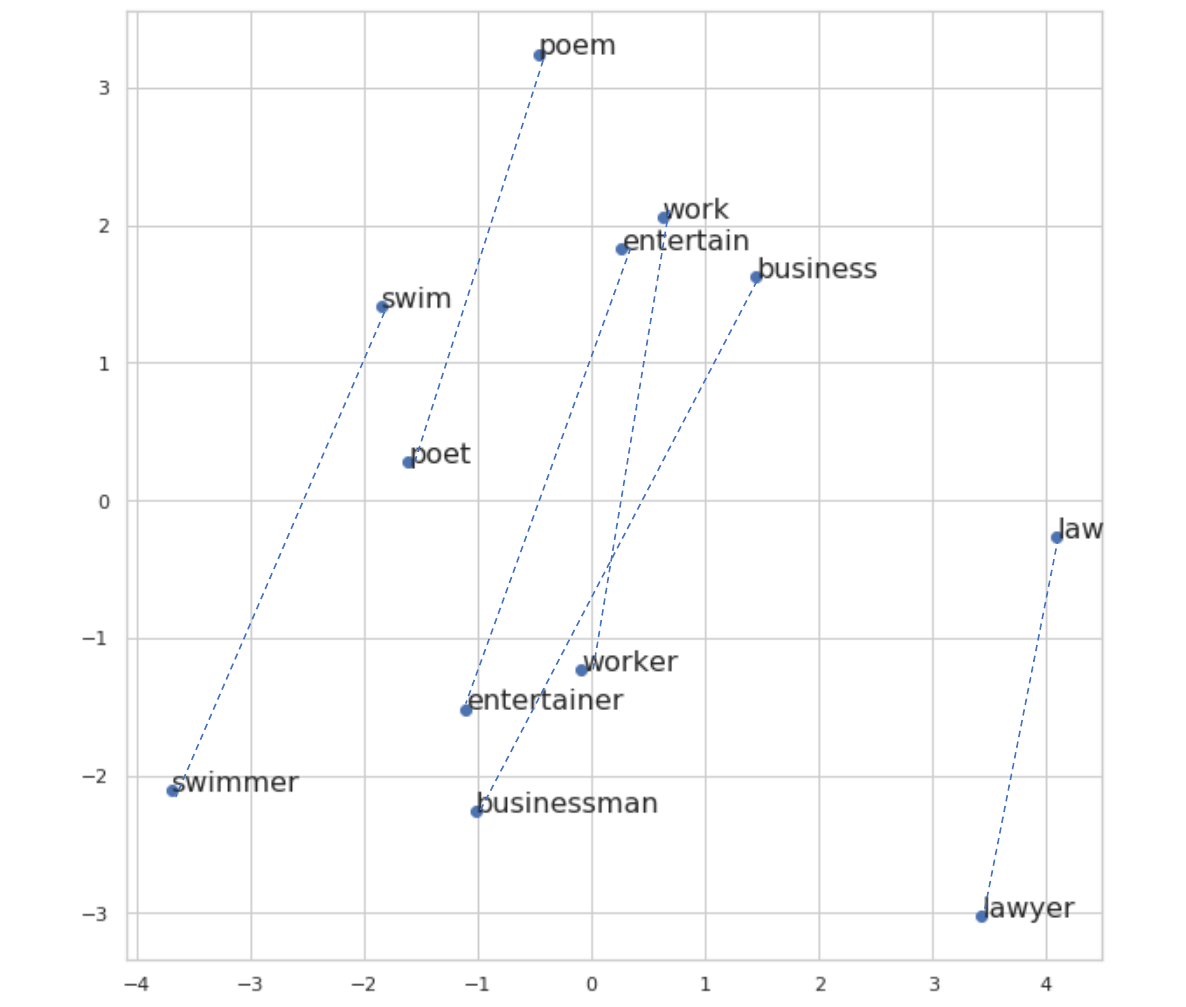}
\end{center}
\caption{PCA visualization of a new word analogy task: ``profession'', which are automatically generated by the ``profession'' word factor.}
\label{fig: pca-profession}
\end{figure}

\noindent {\bf Unclear Factors.} 
% \textcolor{orange}{(BC: This part needs a better title. This doesn't tell me much about what this section describes.)}
While most of the factors have a strong and clear semantic meaning, there are also about less than 10\% of them that we can not identify.
Some of these factors have relatively dense activations that they may activate on more than 10\% of the whole vocabulary while the activations are relatively low. Some of the factors seem to cluster either high or low frequency words regardless of the semantics, e.g. a factor's top-activation words are all rare words. We feel that some of these unidentifiable factors might actually have semantic meaning with more summarization effort and the rest might be due to an optimization choice, e.g. we sampled word embedding in proportion to the words' frequency during optimization, so that high frequency words got more exposure than low frequency ones.
%{\bf factors unable to identify. high/low freqs}

% \begin{figure}[t!]
%     \begin{subfigure}[b]{0.50\linewidth}
%         \includegraphics[width=\linewidth]{images/new-glove-gender-bar.png}
%         \caption{Female factor.}
%         \label{fig: genderbar}
%     \end{subfigure}
%     ~
%      \begin{subfigure}[b]{0.50\linewidth}
%         \includegraphics[width=\linewidth]{images/new-glove-superlative-bar.png}
%         \caption{Superlative factor.}
%         \label{fig: superlativebar}
%     \end{subfigure}
%     \caption{The reliability of the factors.}
% \end{figure}

\noindent {\bf Factor Groups.}
Different factors may correspond to similar semantic meaning and in a particular word vector, they co-activate or only one of them activate. But in general similar factors tend to have a relatively higher co-activation. Based on the co-activation strength, we can cluster word factors in to groups, each provides a more reliable semantic and syntactic meaning detection. In Appendix~\ref{sec:visual_factorgroup}, we show the co-activation patterns in more details.

\subsection{Word Vector Visualization}
\label{sec:visual_wordvector}
\begin{figure*}[htb]
\begin{center}
\includegraphics[width = 0.9\linewidth]{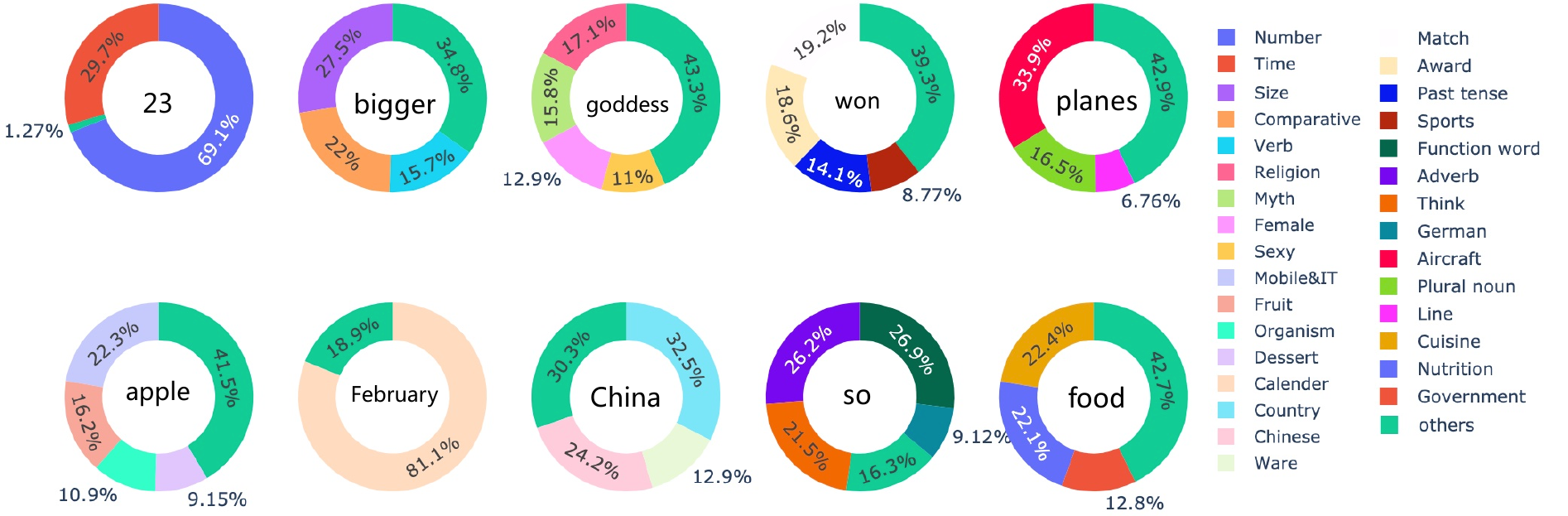}%formula-pie.pdf}
\end{center}
\caption{Word vectors can be decomposed into a sparse linear combination of word factors. Due to a space limit, we only show the major factors and leave the rest as ``others''.}
\label{fig: formula}
\end{figure*}

As a result of sparse coding, every word vector can now be expressed as a linear combination of a limited number of word factors. This makes it possible for us to open up continuous word vectors and see different aspects of meanings through the component factors. In Figure \ref{fig: formula}, we show several word vectors as a combinations of highly activated factors.

% \makeatletter\def\@captype{figure}\makeatother
% \begin{minipage}{.45\textwidth}
% \centering
% \includegraphics[width=\textwidth]{images/apple.png}
% \caption{word vector visualization: apple}
% \label{fig: apple}
% \end{minipage}
% \makeatletter\def\@captype{table}\makeatother
% \begin{minipage}{.45\textwidth}
% \centering
% \begin{tabular}{|c|l|}
% \hline
% \multicolumn{1}{|l|}{factors} & \multicolumn{1}{c|}{top words}                \\ \hline
% \multirow{2}{*}{296}          & 3gs, ipad, jailbroken, iphone, jailbreak,      \\
%                               & iphones, appstore, jailbreaking, icloud, \\
%                               & ibooks, ios, facetime, 4g, ipads, 3g                  \\ \hline
% \multirow{2}{*}{582}          & tangerines, peaches, mangoes, pears, mangos, \\
%                               & papaya, pineapples, apples, quinces, kiwis, \\
%                               & cantaloupe, oranges, seedless, plums, papayas \\ \hline
% \end{tabular}
% \caption{Polysemy opened up}
% \label{tab: top words}
% \end{minipage}

\noindent {\bf Polysemy Separation.} 
Words like ``apple'' contain multiple senses of meanings but are encoded into one continuous vector. By visualization through word factors, different senses are separated. As is shown in figure \ref{fig: formula}, the vector of ``apple'' contains 4 major factors: ``technology'', ``fruit'', ``dessert'' and ``organism''. This combination coincides with our knowledge that ``apple'' is a fruit, a food ingredient, a living creature and a well-known tech company. Another polysemous example is the word ``China'': the presence of factor ``ware'' makes sense as the training corpus is not case sensitive. We further notice the ``country'' factors and the ``Chinese'' factor, which is closely related to specific Chinese nouns such as the names of its provinces and cities. In fact a combination of the ``country'' factor and a ``country-specific'' factor shows up as a common combination in the names of countries.

\noindent {\bf Semantic $+$ Syntactic.}
Besides polysemy, we also find that words are opened up as combinations of both semantic and syntactic factors: ``big'' has both ``size'' and ``comparative'' factor; ``won'' has ``match'', ``award'', ``sports'' and of course ``past tense''. 
% In Appendix~\ref{app:formulas}, we provide many more interesting examples and formulas.

\noindent {\bf Unexpected Meaning.}
Sometimes a word may have an unexpected factor, e.g. in the visualization of ``so'', we find a ``German'' factor, of which all the top activated words are German word pieces 
% \textcolor{red}{(AL:these are word pieces rather than words per se)} 
like ``doch'', ``aber'', ``voll'', ``schön'' and ``ich''. The possible explanation for this is that the training data of the embedding model covers German corpus, and ``so'' is actually also used in German.

\noindent {\bf Word Vectors Manipulations.}
Prior work \cite{mikolov2013linguistic} has demonstrated that linear operations between continuous word vectors can capture linguistic regularities. Now given such factors with clear and strong semantic meanings, it is natural to think of some manipulations. An interesting question is: if we manually add in or subtract out a certain factor from a word vector, would the new word vector be consistent with the semantic relations entailed by the manipulation? To validate this, we manipulate a vector with some factors, and see what is the nearest word in the embedding space. Examples are listed in Table~ \ref{tab:manipulation}. Since the average norm of the word vectors we use is about 7.2, while the factors are of unit norm, we give a constant coefficient of 4 to the factors so that their lengths become comparable but still shorter than the word vectors, therefore can be appropriately regarded as components of word vectors. Results show that both syntactic and semantic meanings including part of speech, gender, sentiment and so on are successfully modified in the desired way. This interesting experiment shows the potential of word factors. Given their explicit meanings, now we are no longer limited to operations between word vectors, but can also conduct operations between word vectors and factors.

% \begin{figure}[htb]
% \begin{center}
% \includegraphics[width=\linewidth]{images/mani.pdf}
% \end{center}
% \caption{factor - vector manipulations.}
% \label{fig: mani}
% \end{figure}

\begin{table}[]
\small
\centering
\caption{Factor-vector manipulations and factor-factor manipulations. The word vectors' average norm is 7.2. The learned word factors all have unit norm.}
\begin{tabular}{|c|c|}
\hline
{\bf Manipulations} & {\bf Nearest Neighbors} \\ \hline
$V_{good} + 4F_{superlative}$  & $V_{best}$  \\ \hline
$V_{king} + 4F_{female}$ & $V_{queen}$ \\ \hline
$V_{Italy}+4F_{national}$ & $V_{Italian}$ \\ \hline
$V_{unwise}+4F_{negative\ prefix}$  & $V_{prudent}$\\ \hline
$V_{hospital}+4F_{vehicle}$  & $V_{ambulance}$\\ \hline
$V_{soldier}-4F_{military}$  & $V_{man}$\\ \hline
$V_{dancers}-4F_{profession}$ & $V_{dance}$\\ \hline
$V_{kindle}-4F_{book}$ & $V_{ipad}$\\ \hline
\end{tabular}
\label{tab:manipulation}
\vspace{-0.1in}
\end{table}

\subsection{Comparison Across Different Models}
\label{sec:visual_models}

\begin{figure}[htb]
\begin{center}
\includegraphics[width=0.8\linewidth]{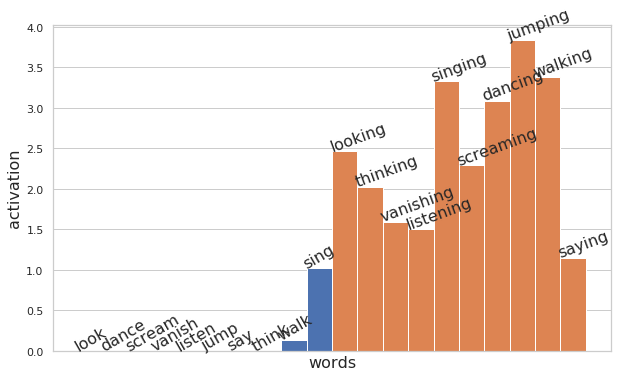}
\end{center}
\caption{The word vector ``sing'' learned by fastText has a high activation on the ``present tense'' factor, due to the subword structure `ing'.}
% \vspace{-0.1in}
\label{fig: sing}

\end{figure}

We conducted experiments with several mainstream word embedding models, including GloVe, Fasttext, Word2vec CBoW and Word2vec skipgram, all of 300 dimension. For GloVe, we download model pretrained on CommonCrawl \cite{pennington2014glove}. For fasttext, we download model pretrained on Wikipedia 2017, UMBC webbase corpus and statmt.org news dataset \cite{Fasttextpretrain}. And for Word2vec models, we trained them on 3B token wikipedia dump \cite{wikidump}.
% \textcolor{orange}{(BC: include citations for these word embedding methods.)}
Although the results are similar between different embedding models, we also notice some interesting differences that can provide understanding of the models. In the fastText embeddings, word ``sing'' has a abnormally high activation on the factor ``present tense'', as is shown in Figure \ref{fig: sing}. This is because the algorithm trains word embeddings based on subword n-grams \cite{bojanowski2017enriching, Joulin2016BagOT}, in this way word ``sing'' is considered as if a word in present tense because it contains a three-gram ``ing''. This shows that despite the advantages of using subword n-grams to embed words, such as tackling out-of-vocabulary issue and encode strong syntactic information, it may also lead to problems. Such a visualization can be used to diagnose and provide insights to improve the existing methods.

\begin{figure}[htb]
\begin{center}
\includegraphics[width=0.9\linewidth]{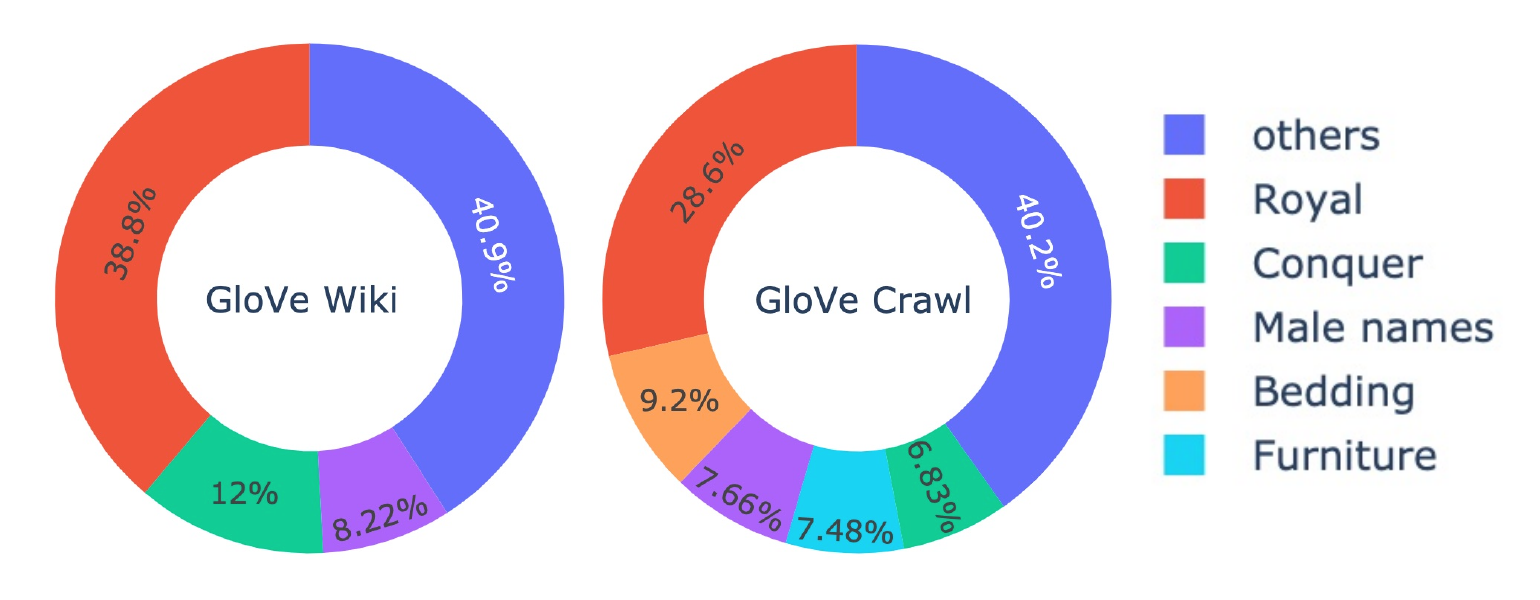}
\end{center}
\caption{Difference between embedding models. The visualization of ``king'' has significant ``bedding'' factor in ``GloVe Crawl'', but it is not found in ``GloVe Wiki''.}
\label{king}
% \vspace{-0.1in}
\end{figure}

% \begin{figure}[htbp]
% \begin{minipage}[t]{0.5\linewidth}
% \centering
% \includegraphics[width=0.9\textwidth]{images/ft-sing-ing-1.png}
% \caption{fastText: word "sing" on present tense factor.}
% \label{fig: sing}
% \end{minipage}%
% \begin{minipage}[t]{0.5\linewidth}
% \centering
% \includegraphics[width=0.9\textwidth]{images/difference-king.pdf}
% \caption{Difference between embeddings.}
% \label{king}
% \end{minipage}
% \end{figure}

% \begin{figure}[h]
% \begin{center}
% \includegraphics[width=\textwidth]{images/ft-sing-ing-1.png}
% \end{center}
% \caption{fastText: word "sing" on present tense factor.}
% \label{fig: mani}
% \end{figure}

There are also differences that reflect bias in different corpus. For example, we compared the pretrained GloVe and a GloVe model trained only on Wikipedia, and refer to them as ``GloVe Crawl'' and ``GloVe Wiki''. As a result, we discovered the factor ``bedding'' in the visualization of vector ``king'' in the ``GloVe Crawl'', while it is missing in ``GloVe Wiki''. The only difference between the two models is the training data. 
%{\bf (TODO: This needs some work .. it's a little confusing without first talking about the datasets with which these models are training on.): The first paragraph in this section.} 
The presence of factor ``bedding'' actually makes sense because ``king'' is frequently used to describe the size of beds and bedclothes. The reason why it is missing in ``GloVe Wiki'' is likely due to the difference in training data. Which is to say such usage of ``king'' is much less frequent in Wikipedia than in CommonCrawl, so it is possible that this aspect of meaning is not significant on Wikipedia.
%{\bf (TODO: only a guess .. we shall compute some statistics if time permits. The arguement is not supported and it's quite shaky. Add: GloVe on wikipedia also no bedding factor, statistics on wikipedia.)} 
In order to verify this, we examined the average co-occurrence statistics in Wikipedia between ``king'' and top 100 activated words of factor ``royal'' and ``bedding''. The fact that the former is more than 30 times larger than the latter supports the assumption that ``king'' appears rather rarely in the ``bedding'' context. Thus a factor of ``bedding'' is hard to get given the corpus from Wikipedia. This comparison shows that the difference in data is captured by embedding models and displayed by our factors.

\section{Improvement in Semantic and Syntactic Tasks with Word Factors}
\label{sec:improve}
% {\bf Every major word embeddings}\\
% \usepackage{multirow}
\begin{table*}[]
\footnotesize
% \vspace{0.1in}
\caption{Performance on word analogy task for different word embedding models}
\centering
\label{tab: perform}
\begin{tabular}{ccccccccc}
\multicolumn{1}{l}{}& \multicolumn{1}{c}{\begin{tabular}[c]{@{}c@{}}W2V \\ sg ori\end{tabular}}
& \multicolumn{1}{c}{\begin{tabular}[c]{@{}c@{}}W2V \\ sg group\end{tabular}}
& \multicolumn{1}{c}{\begin{tabular}[c]{@{}c@{}}Fasttext \\ ori\end{tabular}}
& \multicolumn{1}{c}{\begin{tabular}[c]{@{}c@{}}Fasttext \\ group\end{tabular}}
& \multicolumn{1}{c}{\begin{tabular}[c]{@{}c@{}}W2V \\ cb ori\end{tabular}}
& \multicolumn{1}{c}{\begin{tabular}[c]{@{}c@{}}W2V \\ cb group\end{tabular}}
& \multicolumn{1}{c}{\begin{tabular}[c]{@{}c@{}}GloVe \\ ori\end{tabular}}
& \multicolumn{1}{c}{\begin{tabular}[c]{@{}c@{}}GloVe \\ group\end{tabular}}\\ \hline
0  &  93.87  &  93.87  &  56.92  &  {\bf 58.10}  &  88.34  &  {\bf 89.13}  &  80.04  &  83.99\\ \hline
1  &  89.86  &  {\bf 90.52}  &  40.01  &  {\bf 40.79}  &  87.48  &  {\bf 87.97}  &  79.86  &  {\bf 83.52}\\ \hline
2  &  11.06  &  {\bf 12.02}  &  36.54  &  {\bf 37.26}  &  16.35  &  {\bf 18.27}  &  20.19  &  {\bf 22.84}\\ \hline
3  &  72.15  &  {\bf 75.76}  &  23.89  &  {\bf 24.11}  &  68.22  &  {\bf 68.79}  &  65.46  &  {\bf 66.11}\\ \hline
4  &  86.17  &  {\bf 87.15}  &  89.13  &  {\bf 89.92}  &  89.33  &  {\bf 90.12}  &  96.44  &  {\bf 98.02}\\ \hline
5  &  32.16  &  {\bf 44.56}  &  75.00  &  {\bf 76.61}  &  34.38  &  {\bf 38.81}  &  43.15  &  42.24\\ \hline
6  &  44.46  &  {\bf 47.91}  &  68.97  &  {\bf 71.18}  &  40.39  &  {\bf 43.47}  &  35.10  &  {\bf 42.61}\\ \hline
7  &  83.93  &  {\bf 86.71}  &  97.15  &  97.15  &  89.56  &  {\bf 90.24}  &  87.69  &  {\bf 91.67}\\ \hline
8  &  62.59  &  {\bf 73.86}  &  99.15  &  99.15  &  63.83  &  {\bf 69.79}  &  92.23  &  {\bf 95.55}\\ \hline
9  &  66.38  &  {\bf 87.41}  &  97.73  &  {\bf 100.00}  &  69.98  &  {\bf 86.74}  &  82.77  &  {\bf 96.12}\\ \hline
10  &  90.31  &  {\bf 90.43}  &  85.05  &  {\bf 85.74}  &  88.12  &  {\bf 88.68}  &  69.36  &  {\bf 72.67}\\ \hline
11  &  61.99  &  {\bf 63.08}  &  84.94  &  {\bf 90.71}  &  67.05  &  {\bf 71.60}  &  64.10  &  {\bf 83.72}\\ \hline
12  &  76.43  &  71.55  &  96.70  &  {\bf 97.15}  &  80.26  &  {\bf 82.36}  &  95.95  &  91.44\\ \hline
13  &  71.26  &  {\bf 82.99}  &  95.63  &  {\bf 98.74}  &  63.79  &  {\bf 72.07}  &  67.24  &  {\bf 88.51}\\ \hline
Sem  &  79.53  &  {\bf 81.16}  &  40.81  &  {\bf 41.42}  &  77.22  &  {\bf 77.86}  &  72.84  &  {\bf 75.31}\\ \hline
Syn  &  67.95  &  {\bf 73.47}  &  89.38  &  {\bf 91.19}  &  69.32  &  {\bf 74.01}  &  72.59  &  {\bf 79.80}\\ \hline
Tot  &  72.70  &  {\bf 76.63}  &  74.42  &  {\bf 75.86}  &  72.56  &  {\bf 75.59}  &  72.69  &  {\bf 77.96}\\ \hline
\end{tabular}
\end{table*}

Word analogy task is a classic task for evaluating the quality of word embeddings. Proposed first by \cite{mikolov2013efficient}, it measures the accuracy of answering the question : A is to B as C is to D, such as `king' is to `queen' as `man' is to `woman' and `slow' is to `slower' as 'good' is to 'better'. Given word A, B and C, a model must try to predict D. The conventional approach taken by the word embeddings \cite{mikolov2013efficient, pennington2014glove, bojanowski2017enriching} is a vector arithmetic: calculate $B - A + C$, and find its nearest neighbor in the embedding space as the predicted word. While this approach leads to good results, several failure modes are shown in Table~\ref{tab:errors}. Although vector arithmetic returns a word very close to the ground truth in terms of meaning, it fails to identify the key semantic relationship implied in the question.

\begin{table*}[!h]
% \vspace{0.2in}
\caption{Word analogy performance on new tasks for different word embedding models}
\footnotesize
\centering
\label{tab: newtask}
\begin{tabular}{ccccccccc}
\multicolumn{1}{l}{}& \multicolumn{1}{c}{\begin{tabular}[c]{@{}c@{}}W2V sg \\  ori\end{tabular}}
& \multicolumn{1}{c}{\begin{tabular}[c]{@{}c@{}}W2V sg\\  group\end{tabular}}
& \multicolumn{1}{c}{\begin{tabular}[c]{@{}c@{}}Fasttext \\ ori\end{tabular}}
& \multicolumn{1}{c}{\begin{tabular}[c]{@{}c@{}}Fasttext \\ group\end{tabular}}
& \multicolumn{1}{c}{\begin{tabular}[c]{@{}c@{}}W2V cb \\  ori\end{tabular}}
& \multicolumn{1}{c}{\begin{tabular}[c]{@{}c@{}}W2V cb\\  group\end{tabular}}
& \multicolumn{1}{c}{\begin{tabular}[c]{@{}c@{}}GloVe \\ ori\end{tabular}}
& \multicolumn{1}{c}{\begin{tabular}[c]{@{}c@{}}GloVe \\ group\end{tabular}}\\ \hline
Ideology  &  56.00  &  56.00  &  93.33  &  93.33  &  56.50  &  {\bf 58.67}  &  82.00  &  {\bf 85.50}\\ \hline
Profession  &  27.78  &  {\bf 38.30}  &  65.79  &  {\bf 77.19} &  36.55  &  {\bf 45.03}  &  36.26  &  {\bf 53.80}\\ \hline
Adj-to-verb  &  8.97  &  {\bf 13.46}  &  62.82  &  {\bf 65.38}  &  16.03  &  {\bf 16.67}  &  24.36  &  {\bf 26.28}\\ \hline
Total  &  40.53  &  {\bf 44.44}  &  80.42  &  {\bf 84.34}  &  44.54  &  {\bf 48.45}  &  59.56  &  {\bf 67.21}\\ \hline
\end{tabular}
\end{table*}
\begin{table}[h]

% \vspace{0.2in}
\small
\centering
\caption{A few examples to show the typical errors in word analogy tasks using word vectors.}
\begin{tabular}{|c|c|c|}
\hline
Vector arithmetic                                                                  & Ground truth  & Prediction                       \\ \hline
\begin{tabular}[c]{@{}c@{}}unreasonable - \\ reasonable + competitive\end{tabular} & uncompetitive & \multicolumn{1}{l|}{competition} \\ \hline
worse - bad + cheap                                                                & cheaper       & cheapest                         \\ \hline
greece - athens + beijing                                                           & china         & shanghai                         \\ \hline
\begin{tabular}[c]{@{}c@{}}danced - dancing\\  + enhancing\end{tabular}            & enhanced      & enhance                          \\ \hline
\begin{tabular}[c]{@{}c@{}}wife - husband \\ + policeman\end{tabular}              & policewoman   & policemen                        \\ \hline
\end{tabular}
\label{tab:errors}
\end{table}
However, if semantic meaning captured by the factors are reliable enough, they should be able to identify the semantic difference and therefore correct such mistakes very easily. To do this, we propose a simple factor group selection approach, where we require the predicted word to not only be the closest in the embedding space, but also have an higher activation on the specific factor groups than that of the words in the other category. For example, to be selected as an answer, ``queen'''s activation on the ``female'' factor group must be higher than those male words in the subtask. 
% BC: This description is unclear to me, perhaps describe it with the 'A' 'B' 'C' notation in the previous paragraph.
By simply applying this factor group selection approach, we achieve consistent improvement for every embedding algorithm on almost every subtask. For many subtasks, the improvement is quite significant. Three experiments in Table \ref{tab: perform} get a decreased performance, most likely due to the correct factors are not grouped together during spectral clustering.

Besides validating the previously identified relationships, as mentioned in Section~\ref{sec:visual_wordfactor}, we are also able to find many new ones using the factors, such as ``ideology'' (Figure~\ref{fig: pca-ideo}), ``profession'' (Figure~\ref{fig: pca-profession}) and ``adj-to-verb'' (Figure~\ref{fig: pca-adjverb}). The questions are constructed in the same way as word analogy tasks. Here are examples from each new task:

$V_{\text{collectivist}} - V_{\text{collectivism}} = V_{\text{liberal}} - V_{\text{liberalism}}$

$V_{\text{entertain}} -  V_{\text{entertainer}} = V_{\text{poem}} - V_{\text{poet}}$

$V_{\text{sensational}} - V_{\text{sensationalize}} = V_{\text{marginal}} - V_{\text{marginalize}}$

Performance of each embedding method on the new tasks is shown in Table~\ref{tab: newtask}, which has the same behavior as shown in Table~\ref{tab: perform}. Consistent improvements are obtained once we apply the simple factor group selection method. This further demonstrates that word factors can capture reliable semantic meanings and the phenomenon is not only constraint to the previously proposed ones in the word analogy task.
% BC: were these new tasks manually created? If they were created automatically using the word factors, wouldn't it be expected to better because the new task is generated by the word factors
\vspace{-0.05in}
\section{Discussion}
\label{sec:disscuss}
In this work, we show that dictionary learning can extract elementary factors from continuous word vectors. By using the learned factors, we can meaningfully decompose word vectors and visualize them in many interesting ways. Further, we demonstrate with these factors, we can consistently improve many word embedding methods in word analogy tasks. The word factors may provide an convenient mechanism to develop new word analogy tasks beyond the existing 14 ones. Further examination of existing word embedding models may leads to further improvements.

A fundamental question that remains to be answered is why word factors can be combined in such linear fashion? \cite{levy2014linguistic} provides one possible explanation: with sparse word representation, which explicitly encode each word's context statistics, one can also construct equally good word vectors.
% BC: the previous sentence is not describing the sparse word representation method clearly. This may require a lot more text to describe it better, is it worth it?
If we see a word vector as an explicit statistic based on a word's surrounding context, then this context may fall into sub-categories of words. Our guess is that the learned word factors reflects each of these sub-context categories. This suggests an interesting future direction of our work. A limit of this work is that all the word vectors we visualize are trained from methods which ignore the context of a word used in a specific {\it instance}. Applying dictionary learning to contexualized word vectors \cite{peters2018deep} and attention-based methods \cite{vaswani2017attention, devlin2018bert} is another interesting future direction. In fact, our preliminary results show that the method provided here generalize to different transformer models. Finally, the existence of more elementary meaning than words is a debatable argument in linguistic study. The learned word factors may also provide insights and verification to the sememe thoery \cite{Bazell1966sememe, Niu2017ImprovedWR, Xie2017LexicalSP}. We leave this to the future work.

% \subsubsection*{Acknowledgements}
% We thank Yuhao Zhang, Angli Liu and Ji He for providing many valuable suggestions in the different stages of this work. This work is supported by NSF award 1718991, NSF GRFP Fellowship.

% Use the unnumbered third level heading for the acknowledgements.  All
% acknowledgements go at the end of the paper.

% \subsubsection*{References}

% References follow the acknowledgements.  Use an unnumbered third level
% heading for the references section.  Any choice of citation style is
% acceptable as long as you are consistent.  Please use the same font
% size for references as for the body of the paper---remember that
% references do not count against your page length total.

\renewcommand\bibsection{\subsubsection*{\refname}}
\bibliographystyle{acl_natbib}
\bibliography{AISTATS2020}

% \clearpage
\appendix
\counterwithin{figure}{section}
\counterwithin{table}{section}

\clearpage
\section*{Appendix}

\section{The Word Factor Naming Procedure}
\label{sec:naming}
In this section we illustrate how the factors are named. 

A factor is named based on the common aspect of its top-activation words. Specifically, for every factor, we use the word frequency to weight the factor's  activation on each word, and take the top words that totally contributing 20\% of the total weighted activation. The idea is that a factor should be better represented by words that have strong and obvious activation and show up frequently as well. Usually we get up to 200 words but the number varies from factor to factor. We demonstrate four factors: ``national'', ``mobile\&IT'',  ``superlative'', ``ideology'', among which the first two are semantic factors and the latter two are syntactic factors.
% and with the level of sparsity.

\vspace{0.1in}
\noindent {\bf ``national'' Factor.}
The top 20\% activation of the No.781 factor contains about 80 words. They are enumerated as the following:

\textit{croatian, american, lithuanian, norwegian, vietnamese, chinese, romanian, bulgarian, indonesian, armenian, serbian, turkish, hungarian, korean, malaysian, italian, austrian, portuguese, mexican, macedonian, german, scottish, albanian, cambodian, bosnian, rican, filipino, lebanese, swedish, estonian, irish, venezuelan, dutch, pakistani, haitian, iranian, peruvian, argentine, malay, colombian, danish, ethiopian, australian, european, chilean, brazilian, israeli, japanese, indian, finnish, singaporean, african, british, nigerian, argentinian, belgian, hispanic, french, cypriot, guatemalan, latvian, russian, welsh, algerian, bolivian, egyptian, moroccan, belarusian, jamaican, icelandic, samoan, uruguayan, georgian, ukrainian, jordanian, flemish, muslim, yugoslav, greek, jewish}

By looking into these words, we can easily identify that almost every one of them is related to a specific national meaning, thus this factor can be named ``national''.

\vspace{0.1in}
\noindent {\bf ``mobile\&IT'' Factor.}
The top 20\% activation of the No.296 factor contains about 130 words. They are enumerated as the following:

\textit{ipad, iphone, ios, itunes, apple, android, app, ipod, airplay, 3g, 4s, apps, ipads, htc, tablet, macbook, kindle, galaxy, jailbreak, iphones, netflix, mac, os, touch, nook, skyfire, dock, siri, eris, 4g, thunderbolt, tablets, google, nexus, ipa, barnes, blackberry, sync, devices, hulu, device, ota, amazon, spotify, macworld, retina, wifi, g1, gb, facebook, nfc, syncing, downgrade, protector, iplayer, ipods, 2g, tethered, zune, lte, instagram, s3, kinect, usd, itv, mackintosh, tethering, rooted, tether, shuffle, sansa, garageband, nano, wwdc, smartphones, downgraded, dsi, hotspot, jailbreaking, gps, drm, icloud, smartphone, playbook, casemate, twitter, 3gs, droid, gen., ics, snapchat, multitasking, fuze, stylus, docking, pandora, docks, gadgets, rhapsody, powerbook, tv2, synced, fw, appstore, skype, armband, hd, macbooks, ipo, ssd, evo, aggregator, eyetv, macintosh, g5, folios, steve, sd, gestures, lumia, gen, keynote, shazam, 5g, jellybean, androids, ipcc, cases, magicjack, aria}

By looking into these words, we can easily identify that almost every one of them is related to mobile devices and IT technology, such as apps, brands and etc. Thus we name factor No.296 as ``mobile\&IT''.

Such naming procedure is less subjective if a factor captures syntactic meaning. 

\vspace{0.1in}
\noindent {\bf ``superlative'' Factor.}
For instance, the top 20\% activation of the No.337 factor contains about 70 words:

\textit{strongest, funniest, largest, longest, oldest, fastest, wettest, tallest, heaviest, driest, sexiest, scariest, coldest, hardest, richest, biggest, happiest, smallest, toughest, warmest, most, brightest, loudest, shortest, costliest, coolest, smartest, darkest, slowest, weakest, greatest, lightest, deadliest, thickest, craziest, sunniest, deepest, quickest, busiest, best, cleanest, saddest, worst, ugliest, densest, sweetest, nicest, wealthiest, hottest, weirdest, dumbest, dullest, poorest, highest, bloodiest, prettiest, grandest, safest, meanest, bravest, strangest, catchiest, dirtiest, proudest, cleverest, purest, quietest, fairest, youngest, 
sharpest}

It's clear that this factor captures the ``superlative'' form of different words.

\vspace{0.1in}
\noindent {\bf ``ideology'' Factor.}
Finally, we demonstrate the top 20\% activating about 120 words of the No.674 factor:

\textit{nationalism, liberalism, socialism, individualism, capitalism, communism, fascism, anarchism, materialism, humanism, secularism, feudalism, republicanism, modernism, conservatism, rationalism, imperialism, totalitarianism, militarism, multiculturalism, feminism, marxism, racism, ideology, consumerism, pacifism, modernity, romanticism, utilitarianism, fundamentalism, positivism, democracy, authoritarianism, patriotism, unionism, politics, environmentalism, internationalism, paganism, absolutism, nazism, radicalism, commercialism, pluralism, naturalism, colonialism, protestantism, relativism, idealism, egalitarianism, patriarchy, sexism, spiritualism, libertarianism, regionalism, atheism, mysticism, populism, collectivism, ideologies, pragmatism, universalism, isolationism, anarchy, paternalism, antisemitism, protectionism, federalism, transcendentalism, deism, religiosity, elitism, determinism, neoclassicism, postmodernism, centralism, orthodoxy, empiricism, industrialization, catholicism, puritanism, monasticism, separatism, promoted, realism, classicism, altruism, zionism, nihilism, bolshevism, globalization, sectarianism, progressivism, expressionism, orientalism, morality, modernization, barbarism, christianity, occultism, expansionism, slavery, interventionism, traditionalism, tyranny, monogamy, surrealism, abolitionism, primitivism, hedonism, vegetarianism, historicism, chauvinism, humanitarianism, asceticism, dualism, doctrine, unitarianism, misogyny, extremism}

The idea that it reflects ideology forms of different concepts is quite obvious once we see the words. So the factor summarized as ``ideology''.

% \section{The Details of the Non-negative Sparse Coding Optimization and Spectral Clustering}
\section{The Details of the Non-negative Sparse Coding Optimization}
\label{sec:optimization}
As a convention, all the word vectors used in this word is 300 dimensional and we choose our dictionary to have 1000 word factors.\footnote{We also experimented other settings and they all lead to qualitatively similar result and discussing the difference is beyond the scope of this work.} To learn the word factors, we use a typical iterative optimization procedure:
\begin{equation}
\min\limits_{A} \tfrac{1}{2} \| X - \Phi A\|_{F}^{2} + \lambda\sum_{i}{ \|\alpha_i\|_{1}},\ \text{s.t.}\ \alpha_i \succeq 0,
\label{appequ:sparse_coding}
\end{equation}
\begin{equation}
\min\limits_{\Phi} \tfrac{1}{2} \| X - \Phi A\|_{F}^{2},\ \|\Phi_j\|_2 \leq 1.
\label{appequ:dictionary_update}
\end{equation}

These two optimizations are both convex, we solve them iteratively to learn the word factors: In practice, we use minibatches contains 100 word vectors as $X$. Optimization~\ref{appequ:sparse_coding} can converge in 500 steps using the FISTA algorithm. We experimented with different $\lambda$ values from 0.3 to 1, and choose $\lambda=0.5$ to give results presented in this paper. Once the sparse coefficients have been inferred, we update our dictionary $\Phi$ based on Optimization~\ref{appequ:dictionary_update} by one step using an approximate second-order method, where the Hessian is approximated by its diagonal to achieve an efficient inverse \cite{duchi2011adaptive}. The second-order parameter update method usually leads to much faster convergence of the word factors. Empirically, we train 200k steps and it takes about 2-3 hours on a Nvidia 1080 Ti GPU.

% To better perform a spectral clustering, we first make the normalized covariance matrix $W$ described in section~\ref{sec:learning} by selecting k largest values in each row of $W$:
% \begin{equation}
%     W_{sp} = f_k(W, dim=0)
% \end{equation}
% Where $f_k$ stands for keeping the k largest values unchanged in the given dimension, while set all the other entities to 0. Then we obtain a symmetric sparse matrix:
% \begin{equation}
%     W_{adj} = W_{sp}+W_{sp}^T
% \end{equation}
% $W_{adj} \in {\rm I\!R}^{d\times d}$ is the  adjacency matrix used in spectral clustering \cite{ng2002spectral,von2007tutorial}. We use the implementation of the algorithm in Scipy\cite{scipy}.
% Using the notation from \cite{von2007tutorial}, we first compute a normalized Laplacian matrix $L_{sym}$:
% \begin{equation}
%     L_{sym} = I - D^{-1/2}W_{adj}D^{-1/2}
% \end{equation}
% where $D$ is a diagonal matrix with the sum of each row of the symmetric $W_{adj}$ on its diagonal. Suppose we set the number of clusters to k, then the first k eigenvectors of $L_{sym}$ form the columns of matrix $V \in {\rm I\!R^{d \times k}}$:
% \begin{equation}
%     V = [v_1, v_2, ..., v_k] 
% \end{equation}
% And we normalize each row of V to get a new matrix $U$:
% \begin{equation}
%     U = [u_1, u_2, ..., u_d]^T
% \end{equation}
% where each row $u_i = V_{i,:}/ \Vert V_{i,:} \Vert_2$. $\Vert V_{i,:} \Vert_2$ indicates the L2 norm of the \textit{i}th row of V. Finally, a k-means algorithm is run on the $d$ rows of U to get the clusters. 

\section{The New Word Analogy Tasks Generated}
\label{sec:newanalogy}
\begin{figure}[htb]
\begin{center}
\includegraphics[width=0.8\linewidth]{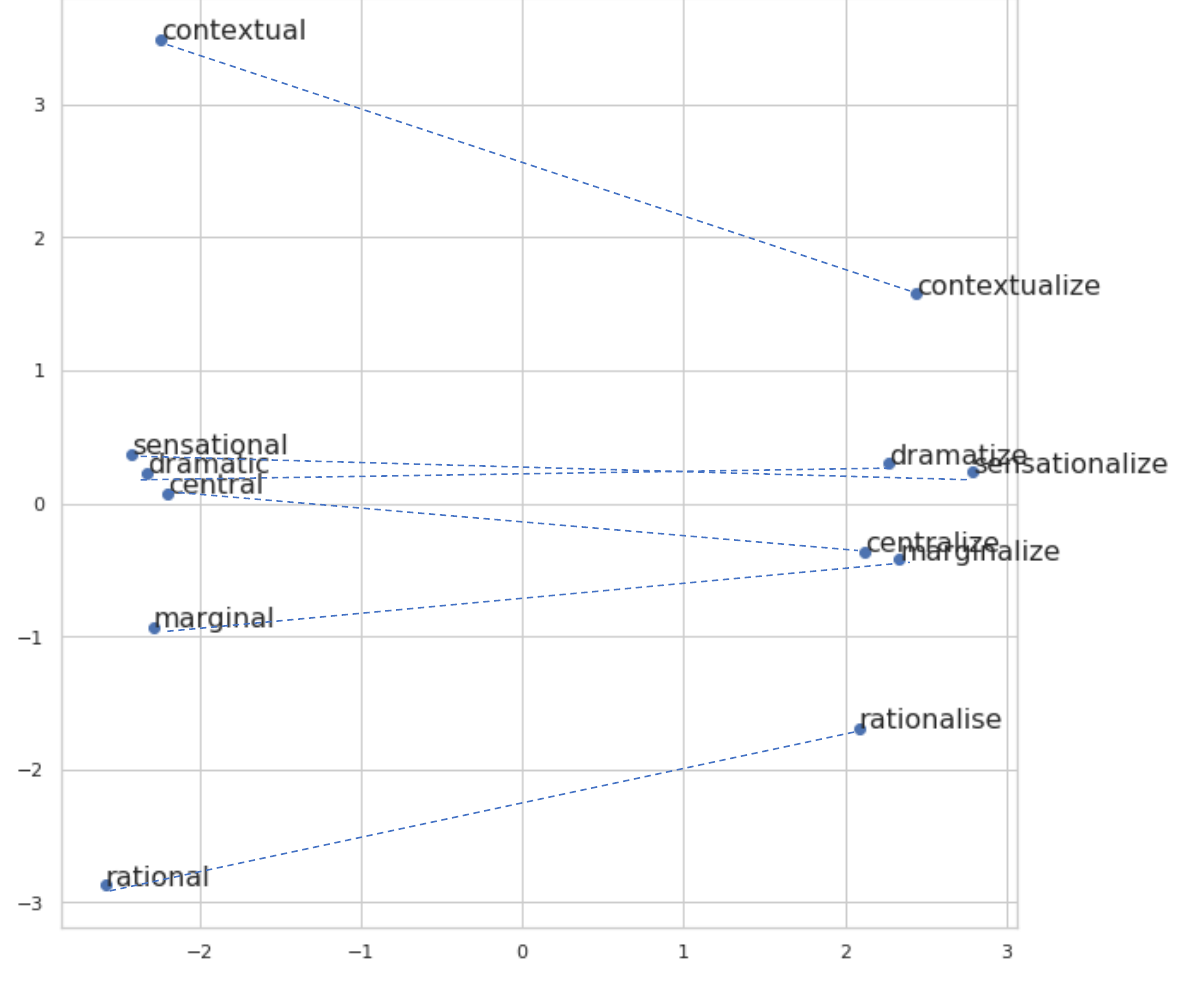}
\end{center}
\caption{PCA of the generated adj-to-verb examples.}
\label{fig: pca-adjverb}
\end{figure}

\begin{figure}[htb]
\begin{center}
\includegraphics[width=0.8\linewidth]{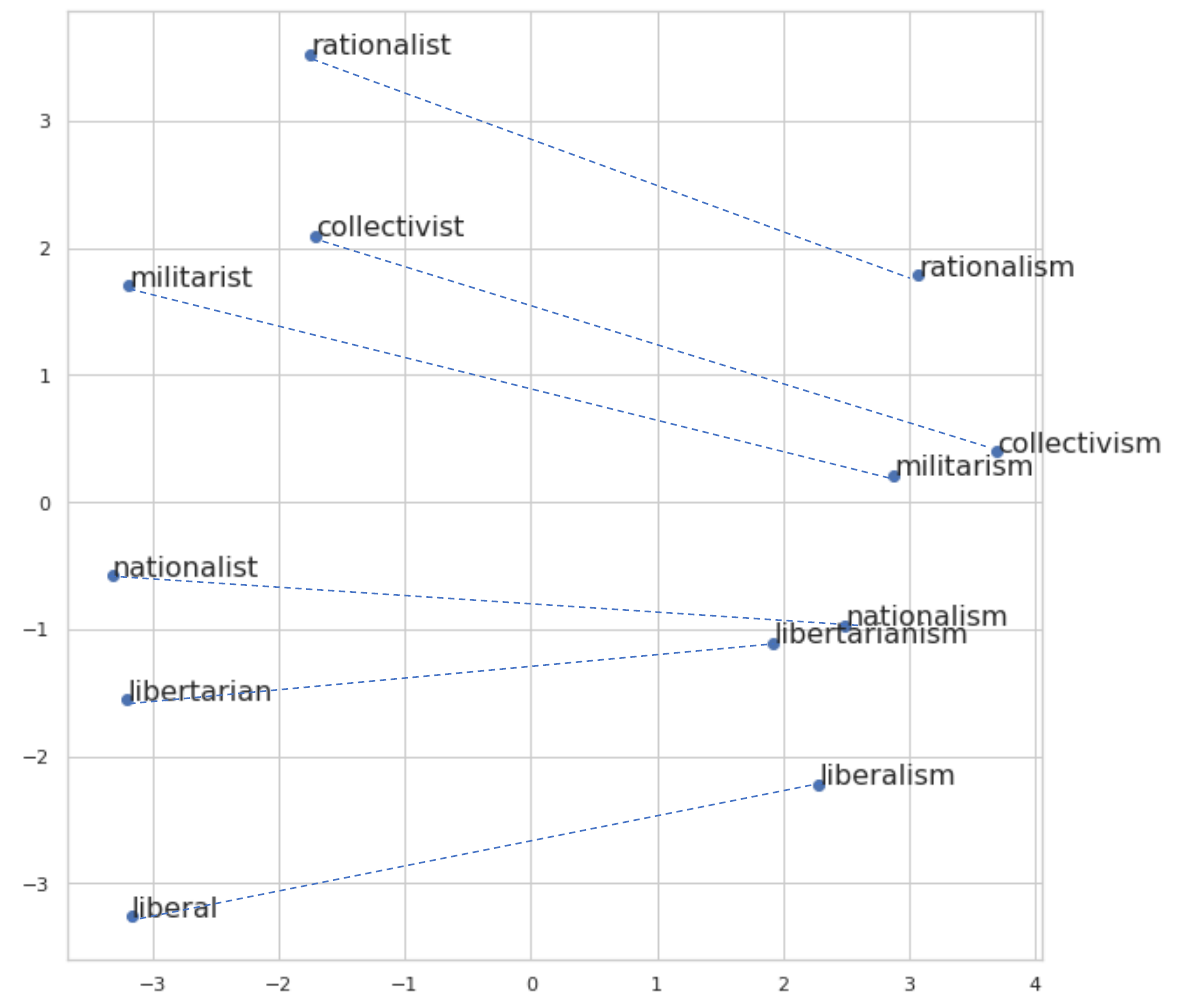}
\end{center}
\caption{PCA visualization of a new word analogy task: ``ideology'', which are automatically generated by the ``ideology'' word factor.}
\label{fig: pca-ideo}
% \vspace{-0.2in}
\end{figure}

In this section, we would like to demonstrate further that the word factors are more elementary structures than word vectors with Figure~\ref{fig: pca-adjverb} and Figure~\ref{fig: pca-ideo}.

\section{Factor Group Co-activation}
\label{sec:visual_factorgroup}
\begin{figure}
\begin{center}
\includegraphics[width=\linewidth]{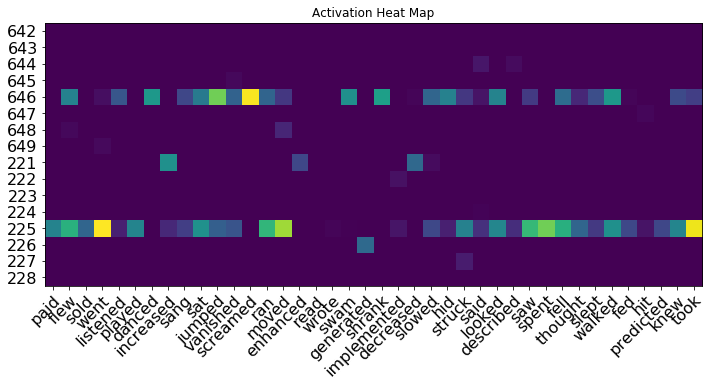}
\end{center}
\caption{This figure shows the co-activation pattern of factors in the ``past tense'' factor group.}
\label{fig: fig:group1}
% \vspace{-0.1in}
\end{figure}

\begin{figure}
% \vspace{-0.1in}
\begin{center}
\includegraphics[width=\linewidth]{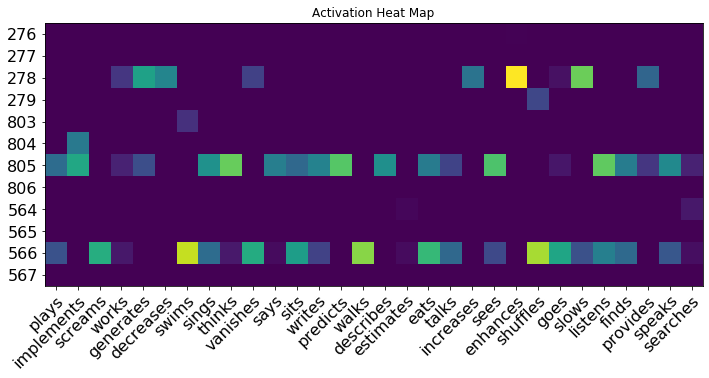}
\end{center}
\caption{This figure shows the co-activation pattern of factors in the ``singular form'' factor group.}
\label{fig: fig:group2}
% \vspace{-0.5in}
\end{figure}

In Figure~\ref{fig: fig:group1} and \ref{fig: fig:group2} we use heat maps to visualize the activations of factors within a group. A heat map shows a fraction of the activation matrix $A$ in Equation~\ref{eqn:sparse_code_matrix}, with each row corresponds to a factor, each column to a word. Therefore, a bright block indicates a high activation on the given word and the dark background means 0 values. It is very clear that factors within a group are often activated together on the same words, forming parallel bright bands across the heat maps.

\clearpage
\onecolumn
\section{Word Factors and Relatedness}
% in word factor visualization section
\begin{table*}[!htbp]
\footnotesize
\centering
\begin{tabular}{|c|l|c|l|}
\hline
\bf{Factor}           & \multicolumn{1}{c|}{\bf{Top-activation Words}}                                                                                                                                                                                                        & \bf{Top1 Word}                                                & \multicolumn{1}{c|}{\bf{Nearest Neighbors}}                                                                                                                                                                                                                                                                \\ \hline
``medical''                                                 & \begin{tabular}[c]{@{}c@{}}hospital, medical, physician, \\ physicians, nurse, doctor, \\ hospitals, doctors, nurses, \\ patient,nursing, medicine, care\end{tabular}                      & \begin{tabular}[c]{@{}c@{}}physician\\ 0.75\end{tabular} & \begin{tabular}[c]{@{}c@{}}diagnoses, anesthetist, occupational,\\ reimbursement, treatment, consultant, \\ consultation, practitioners, consults, \\ herbalist, counselor, psychiatry\end{tabular}                    \\ \hline
``vehicle''                                                 & \begin{tabular}[c]{@{}c@{}}vehicles, vehicle, driving, drivers, \\ cars, car, driver, buses, truck, \\ trucks, taxi, parked, automobile, \\ fleet, bus, taxis, passenger, van\end{tabular} & \begin{tabular}[c]{@{}c@{}}vehicles\\ 0.76\end{tabular}  & \begin{tabular}[c]{@{}c@{}}roadside, lorry, equipped, wagon, \\ automaker, hauling, minibuses, \\ automakers, wheel, motoring, wheels, \\ freight, leasing, planes, transit, tanks\end{tabular}                        \\ \hline
``ware''                                                    & \begin{tabular}[c]{@{}c@{}}pottery, bowl, bowls, porcelain, \\ ware, vase, teapot, china, saucer, \\ denby, vases, saucers, ceramic,\\ glass, plates, earthenware, pitcher\end{tabular}    & \begin{tabular}[c]{@{}c@{}}stoneware\\ 0.75\end{tabular} & \begin{tabular}[c]{@{}c@{}}saucers, lustre, ladle, pot, sherd, \\ dish, pans, denby, talavera, stainless, \\ urn, kilns, wares, burnished, earthen, \\ casseroles, dishes, glass, platters\end{tabular}                \\ \hline
\begin{tabular}[c]{@{}c@{}}``mobile \\ \& IT''\end{tabular} & \begin{tabular}[c]{@{}c@{}}ipad, iphone, ios, itunes, apple, \\ android, app, ipod, airplay, 3g, \\ 4s, apps, ipads, htc, tabletgalaxy, \\ jailbreak, iphones, netflix, mac\end{tabular}   & \begin{tabular}[c]{@{}c@{}}ipad\\ 0.80\end{tabular}      & \begin{tabular}[c]{@{}c@{}}firefox, sync, ericsson, tethering, hack, \\ macbooks, symbian, toshiba, ubuntu, \\ galaxy, sim, jailbreaking, spotify, cnet, \\ macworld, lenovo, dsi, gadgets, youtube\end{tabular}       \\ \hline
``superlative''                                             & \begin{tabular}[c]{@{}c@{}}strongest, funniest, largest, \\ longest, oldest, fastest, wettest,\\ tallest, heaviest, driest, sexiest, \\ scariest, coldest, hardest\end{tabular}            & \begin{tabular}[c]{@{}c@{}}meanest\\ 0.71\end{tabular}   & \begin{tabular}[c]{@{}c@{}}straightest, unluckiest, reddest, flattest, \\ hippest, brutish, funniest, contemptible, \\ naughtiest, mildest, fanciest, brainiest, \\ severest, plainest, harshest, fittest\end{tabular} \\ \hline
``country''                                                 & \begin{tabular}[c]{@{}c@{}}venezuela, germany, paraguay, \\ uruguay, norway, russia, lithuania, \\ ecuador, netherlands,estonia, \\ korea, brazil, argentina, albania\end{tabular}         & \begin{tabular}[c]{@{}c@{}}venezuela\\ 0.77\end{tabular} & \begin{tabular}[c]{@{}c@{}}yemen, nacional, belarus, jalisco, libya, \\ ciudad, fiji, haiti, moldova, europe, \\ tunisia, bulgaria, iran, chilean, trujillo, \\ croatia, acapulco, grenada, romania\end{tabular}       \\ \hline
``bedding''                                                 & \begin{tabular}[c]{@{}c@{}}mattress, pillow, bed, mattresses, \\ beds, pillows, queen, ottoman, \\ simmons, cushion, bedding, \\ topper, foam, plush, sleeper, sofa\end{tabular}           & \begin{tabular}[c]{@{}c@{}}mattress\\ 0.73\end{tabular}  & \begin{tabular}[c]{@{}c@{}}mats, chaise, polyurethane, slumber, \\ dryer, twin, orthopedic, flannel, \\ cushioning, stroller, sleepers, \\ cotton, wicker, diapers, snug, drawers\end{tabular}   \\ \hline
``royal''                                                   & \begin{tabular}[c]{@{}c@{}}king, royal, throne, prince,\\ monarch, emperor, duke, queen, \\ reign, coronation, kings, empress, \\ regent, dynasty, palace, monarchs\end{tabular}           & \begin{tabular}[c]{@{}c@{}}monarch\\ 0.72\end{tabular}   & \begin{tabular}[c]{@{}c@{}}highness, decreed, noble, reigning, \\ princess, patriarch, uncrowned, \\ aristocrats, rightful, kingdoms, \\ parliament, overlord, papal, barons\end{tabular}                              \\ \hline
``fruit''                                                   & \begin{tabular}[c]{@{}c@{}}fruit, fruits, pears, oranges, \\ apples, peaches, grapes, apple, \\ ripe, plums, bananas, mandarin, \\ grapefruit, peach, berries, tomatoes\end{tabular}       & \begin{tabular}[c]{@{}c@{}}peaches\\ 0.80\end{tabular}   & \begin{tabular}[c]{@{}c@{}}quinces, onion, pickles, chestnuts, \\ eggplant, honey, sliced, clementines, \\ pumpkin, custard, okra,eggplants, \\ stewed, pudding, avocado, kiwis\end{tabular}                           \\ \hline
``Chinese''                                                 & \begin{tabular}[c]{@{}c@{}}china, fujian, zhejiang, guangdong, \\ hangzhou, shandong, shanghai, \\ qingdao, beijing, chongqing, \\ guangzhou, sichuan, jiangsu\end{tabular}                & \begin{tabular}[c]{@{}c@{}}jiangsu\\ 0.78\end{tabular}   & \begin{tabular}[c]{@{}c@{}}anyang, basf, hongkong, asean, \\ saitama, dhaka, turkestan, zhou, \\ huangpu, zhonghua, yangtze, kunming, \\ ludhiana, kaohsiung, taipei, sumitomo\end{tabular}                     \\ \hline
``national''                                                & \begin{tabular}[c]{@{}c@{}}croatian, american, lithuanian, \\ norwegian, vietnamese, chinese, \\ romanian, bulgarian, indonesian, \\ armenian serbian, turkish\end{tabular}                & \begin{tabular}[c]{@{}c@{}}estonian\\ 0.73\end{tabular}  & \begin{tabular}[c]{@{}c@{}}czechs, flemish, bengali, cebuano, \\ visayan, argentine, singaporean, \\ dutch, malaysian, bulgarians, ethiopian, \\ moldavian, norwegians,slovenia, algerian\end{tabular}                 \\ \hline
``female''                                                  & \begin{tabular}[c]{@{}c@{}}her, queen, herself, she, actress, \\ feminist, heroine, princess, \\ empress, sisters, woman, dowager, \\ lady, sister, mother, goddess\end{tabular}           & \begin{tabular}[c]{@{}c@{}}dowager\\ 0.56\end{tabular}   & \begin{tabular}[c]{@{}c@{}}murderess, snobbish, corpulent, \\ suffragette, bedchamber, shepherdess, \\ gentility, courtier, widow, curtsied, \\ equerry, lordly, ereshkigal, almoner\end{tabular}                      \\ \hline
\end{tabular}
    \caption{Word Factors and Relatedness. In this table we show a set of factors and their top-activation words from Table \ref{tab:factor_words}, together with the nearest words to each factor in the embedding space and their nearest neighbors. It serves as a comparison between factors and word vectors to show that the factor we discovered are not exactly the words already existed.}
    \label{tab:word_words}
\end{table*}
\end{document}